\documentclass[twoside]{article}

\usepackage[accepted]{aistats2022}

\usepackage[shortlabels]{enumitem}
\usepackage{amsmath,amssymb,amsthm,stmaryrd,hyperref,algorithm}
\usepackage{graphicx}

\usepackage[
    backend=biber,
    style=bwl-FU,
    url = false,
    doi = false,
]{biblatex}

\usepackage{caption}
\usepackage{subcaption}
\usepackage[normalem]{ulem}

\usepackage[noend]{algpseudocode}

\newcommand{\E}{\mathbb{E}}
\newcommand{\mc}{\mathcal}
\newcommand{\Tr}{\mathtt{Tr}}
\newcommand{\D}{\mathcal{D}}
\newcommand{\W}{\mathcal{W}}
\newcommand{\eps}{\varepsilon}
\newcommand{\ST}{\mathrm{S}}
\newcommand{\OPT}{\mathrm{OPT}}
\newcommand{\norm}[1]{\left\lVert #1 \right\rVert_2}

\newtheorem{thm}{Theorem}
\newtheorem{lemma}{Lemma}
\newtheorem{claim}{Claim}

\newtheorem{defn}{Definition} 

\newtheorem{ex}{Example}

\newtheorem{assumption}{Assumption}

\newcommand{\Sec}[1]{\hyperref[sec:#1]{Section\,\ref*{sec:#1}}} %
\newcommand{\Obs}[1]{\hyperref[obs:#1]{Observation~\ref*{obs:#1}}} %
\newcommand{\Eqn}[1]{\hyperref[eq:#1]{Eq. (\ref*{eq:#1})}} %
\newcommand{\Fig}[1]{\hyperref[fig:#1]{Figure\,\ref*{fig:#1}}} %
\newcommand{\Tab}[1]{\hyperref[tab:#1]{Table\,\ref*{tab:#1}}} %
\newcommand{\Thm}[1]{\hyperref[thm:#1]{Theorem\,\ref*{thm:#1}}} %
\newcommand{\Fact}[1]{\hyperref[fact:#1]{Fact\,\ref*{fact:#1}}} %
\newcommand{\Lem}[1]{\hyperref[lem:#1]{Lemma\,\ref*{lem:#1}}} %
\newcommand{\Prop}[1]{\hyperref[prop:#1]{Prop.~\ref*{prop:#1}}} %
\newcommand{\Cor}[1]{\hyperref[cor:#1]{Corollary~\ref*{cor:#1}}} %
\newcommand{\Conj}[1]{\hyperref[conj:#1]{Conjecture~\ref*{conj:#1}}} %
\newcommand{\Def}[1]{\hyperref[def:#1]{Definition~\ref*{def:#1}}} %
\newcommand{\Alg}[1]{\hyperref[alg:#1]{Algorithm~\ref*{alg:#1}}} %
\newcommand{\Pro}[1]{\hyperref[pro:#1]{Procedure~\ref*{pro:#1}}} %
\newcommand{\Ex}[1]{\hyperref[ex:#1]{Example~\ref*{ex:#1}}} %
\newcommand{\Clm}[1]{\hyperref[clm:#1]{Claim~\ref*{clm:#1}}} %
\newcommand{\Stp}[1]{\hyperref[step:#1]{Step~\ref*{step:#1}}}
\newcommand{\Ch}[1]{\hyperref[chap:#1]{Chapter~\ref*{chap:#1}}}
\newcommand{\Assumption}[1]{\hyperref[assumption:#1]{Assumption~\ref*{assumption:#1}}}

\algnewcommand\algorithmicinput{\textbf{INPUT:}}
\algnewcommand\INPUT{\item[\algorithmicinput]}
\algnewcommand\algorithmicoutput{\textbf{OUTPUT:}}
\algnewcommand\OUTPUT{\item[\algorithmicoutput]}

\graphicspath{ {figures/} }

\usepackage{xcolor}

\usepackage[symbol]{footmisc}

\begin{document}

\runningauthor{Gavin Brown, Shlomi Hod, Iden Kalemaj}

\twocolumn[

\aistatstitle{Performative Prediction in a Stateful World}

\aistatsauthor{ Gavin Brown$^*$ \And Shlomi Hod$^*$ \And  Iden Kalemaj$^*$ }

\aistatsaddress{ Boston University \And Boston University \\ $^*$equal contribution \And Boston University }]

\begin{abstract}
  Deployed supervised machine learning models make predictions that interact with and influence the world. This phenomenon is called \emph{performative prediction} by Perdomo et al.~(ICML 2020). It is an ongoing challenge to understand the influence of such predictions as well as design tools so as to control that influence. We propose a theoretical framework where the response of a target population to the deployed classifier is modeled as a function of the classifier and the current state (distribution) of the population. We show necessary and sufficient conditions for convergence to an equilibrium of two retraining algorithms, \emph{repeated risk minimization} and a lazier variant. Furthermore, convergence is near an optimal classifier. We thus generalize results of Perdomo et al.,~whose performativity framework does not assume any dependence on the state of the target population. A particular phenomenon captured by our model is that of distinct groups that acquire information and resources at different rates to be able to respond to the latest deployed classifier. We study this phenomenon theoretically and empirically. 
\end{abstract}

\section{INTRODUCTION}

 Supervised learning is widely used to train classifiers that aid institutions in decision-making: will a loan applicant default? Will a user respond well to certain recommendations? Will a candidate perform well in this job?

 Several studies and examples demonstrate that such predictions can influence the behavior of the target population that they try to predict, e.g., \cite{camacho2011manipulation, o2016weapons, ribeiro2020auditing}. 
 Loan applicants strategically manipulate credit card usage to appear more creditworthy, job applicants tailor their resumes to resume-parsing algorithms, and user preferences on a platform shift as they interact with recommended items. 
 It is an ongoing challenge to both understand and address the influence of such predictions.

 In their recent paper ``Performative Prediction,'' Perdomo, Zrnic, Mendler-D{\"u}nner, and Hardt (2020) term such predictions \emph{performative}. They establish a theoretical framework for analyzing performativity in supervised learning and propose \emph{repeated risk minimization} as a strategy that institutions can apply in hopes of converging to an equilibrium. The equilibrium is a classifier that is optimal for the distribution it induces. \cite{perdomo2020} model the response of the target population via a deterministic function $\D$ of the published classifier $\theta$. The distribution $\D(\theta)$ induced by a classifier $\theta$ is unaffected by previously deployed classifiers. 
 However, in practice the environment may depend heavily on the history of classifiers deployed by an institution.

 Consider the following example: individuals applying for loans manipulate their features to receive favorable results from a bank.
 These actions depend not only on the bank's classifier, but also on the previous feature state of the individual.
 As the bank updates its criteria for creditworthiness, new features may become important.
 Furthermore, when a new classifier is published, different groups in the target population acquire information and adapt their behavior at different rates, so that repeated application of the same classification model may result in continued distribution shifts. We propose a performativity framework that allows for such history dependence.

\subsection{Our Framework}

We cast the phenomenon of performativity in repeated decision-making as an online learning game \parencite{shalev_schwartz}. 
At round $t$, the institution chooses a classifier $\theta_t$ to publish. 
In response, the adversary picks a distribution $d_t$ over labeled samples. The institution then suffers loss $\E_{z\sim d_t}[\ell(z;\theta_t)]$ for some fixed loss function $\ell$.
Per convention, we also refer to loss as ``risk.''
For now, we set aside finite-sample issues and assume that the institution observes the distribution directly.

Standard online learning assumes that the adversary may be malicious and pick whichever distribution causes the greatest loss.
To model state and performativity, we propose a weaker adversary that responds according to a \textit{transition map} $\Tr(;)$, mapping classifier-distribution pairs to distributions. 
The transition map is fixed but a priori unknown to the institution.
If the institution plays $\theta$ and the previous distribution played by the adversary was $d$, the adversary responds with 
\[d' = \Tr(d; \theta).
\]
We denote by $\theta_1, \theta_2, \dots$ the classifiers played by the institution, and by $d_1, d_2, \dots$ the distributions played by the adversary.  

We call our framework \emph{stateful}, since it incorporates the current distribution of the target population in the performative response map $\Tr(;)$, thus possibly preserving information about the state of the population and the history of classifiers played by the institution. Our framework generalizes the \emph{stateless} framework of \cite{perdomo2020}, whose implicit ``transition map'' depends only on the current classifier $\theta$.

To illustrate our framework, we provide three theoretical examples that are special instances of the model and might be of independent interest. \Ex{kGRS} and \Ex{GDR} capture the particular phenomenon of individuals who act strategically but with outdated information. They provide a starting point for the study of the disparate effects of performativity. We study \Ex{kGRS} both theoretically and empirically in \Sec{kGRS} and \Sec{simulation}, respectively.

\begin{ex}[$k$ Groups Respond Slowly]\label{ex:kGRS}
    \emph{
    Assume there is a deterministic ``strategic response function'' $\D(\theta)$ unknown to the institution.
    In response to $\theta_t$, the distribution $d_t$ is a uniform mixture of $k$ distributions $(d_t^{|j})_{j \in [k]}$, one for each group, where
    \begin{equation}
        d_t^{|j} = \begin{cases}
            \D(\theta_{t-j+1}) & \text{if $t-j+1\ge 1$} \\
            d_0^{|j} & \text{otherwise}
        \end{cases}.
    \end{equation}
    Here $d_0^{|j}$ is the initial distribution for group $j$.
    Thus, Group 1 adjusts strategically to the current classifier, while groups with higher indices react to correspondingly older classifiers, modeling a setting where distinct groups receive information at different rates.
    When $k=1$ this is the setting of \cite{perdomo2020}. }
\end{ex}

\begin{ex}[Geometric Decay Response]\label{ex:GDR}\emph{
    As in \Ex{kGRS}, assume that there is a fixed $\D(\theta)$.
    The adversary plays a mixture over past responses.
    For $\delta \in [0,1]$, define
    \begin{align*}
        \Tr(d_{t-1}; \theta_t) = (1-\delta) d_{t-1} + \delta \cdot \D(\theta_t).
    \end{align*}
    The mixture coefficients in the current distribution decay geometrically across older responses.
    When $\delta=1$ this is the setting of \cite{perdomo2020}.}
\end{ex}

\begin{ex}[Markov Transitions]\label{ex:markov}
    \emph{To each classifier $\theta$, associate a stochastic matrix $A_\theta$.
    The transition map is defined as $\Tr(d; \theta) = A_\theta d$.}
\end{ex}

\subsection{Our Results}

Our goal is to devise a strategy for the institution that converges towards an approximately optimal distribution-classifier pair. \cite{perdomo2020} propose the strategy of \emph{repeated risk minimization} (RRM) where, at every round, the institution chooses the classifier that minimizes loss on the last distribution played by the adversary:
\begin{equation*}
\theta_{t+1} =  \underset{\theta}{\mathrm{argmin}} \underset{z \sim d_t}{\E} \ell(z;\theta).
\label{update_of_theta}
\end{equation*}
It is a natural strategy, akin to retraining heuristics used in practice to deal with different kinds of distribution shifts. 
\cite{perdomo2020} analyze RRM in the stateless framework and show that, under convexity and Lipschitz assumptions, it will converge to a near-optimal classifier. 

In our setting, when the history of previous classifiers can influence the distribution, it is not clear if such an iterative retraining procedure will succeed.
In  \Ex{rrm_may_not_converge}, we demonstrate that there are parameter settings for which RRM converges in the stateless model but not in the stateful one.

To control the extent of the performative response, we follow the approach of \cite{perdomo2020}~and impose a Lipschitz requirement on the transition map. It ensures that small changes in the distribution or classifier yield only small changes in the updated distribution. Let $\Theta$ denote the set of classifiers, which we assume is a closed convex subset of $\mathbb{R}^d$, and let $\Delta(\mc{Z})$ be the space of distributions over examples. 

\begin{defn}[$\eps$-joint sensitivity] 
	The transition map $\Tr(;)$ is $\eps$-jointly sensitive if, for all $\theta, \theta' \in \Theta$ and $d, d' \in \Delta(\mc{Z})$,
	\begin{equation*}
	\mc{W}_1(\Tr(d; \theta), \Tr(d'; \theta')) \leq  \eps \mc{W}_1(d, d') +  \eps \lVert \theta - \theta'\rVert_2,
	\end{equation*}
	where $\mc{W}_1$ denotes the Wasserstein-1 distance between distributions. 
\end{defn}

In our model, even repeated deployment of the same classifier can cause ``thrashing'' behavior, e.g., alternating between two distributions. Therefore, we focus on scenarios where the Lipschitz parameter satisfies $\eps < 1$. In this case, the map $\Tr(\cdot; \theta):\Delta(\mathcal{Z})\to \Delta(\mathcal{Z})$ is contractive, and repeated application of the same classifier causes the induced distributions to converge to a \emph{fixed point} that depends only on $\theta$. The concept of a fixed point distribution for every classifier is a key aspect of our framework and results. Intuitively, this models behavior where the environment will eventually settle on a single response to the institution's classifier. 

In the setting where $\eps < 1$, we devise algorithms that converge to an equilibrium pair: a fixed point distribution and a classifier that achieves minimum loss on this distribution. Delayed RRM is a first attempt at an algorithm that converges to an equilibrium pair. It repeats the following: repeatedly deploy the same classifier until we approach a fixed point distribution and only then retrain a classifier that minimizes risk on the current distribution. The goal of Delayed RRM is to overcome the stateful aspect of the population's response by only training on distributions that are close to fixed point distributions. However, it suffers the delay of having to deploy the same classifier for multiple rounds.

\begin{thm}[Informal, see \Thm{delayed_rrm} in Supplement]\label{thm:delayed_rrm_informal}
    If the loss function $\ell(z;\theta)$ is smooth and strongly convex and the transition map $\Tr(d;\theta)$ is Lipschitz in both arguments, then Delayed RRM converges to an equilibrium distribution-classifier pair, coming within distance $\delta$ in $O(\log^2 1/\delta)$ rounds.
\end{thm}

As our main result, we show that this delay in retraining is not necessary. We show sufficient conditions, very similar to those in \Thm{delayed_rrm_informal}, under which RRM converges to an equilibrium pair. The rate of convergence is much faster than Delayed RRM. 

\begin{thm}[Informal, see \Thm{iterative_risk_min}]
\label{thm:informal_1}
If the loss function $\ell(z;\theta)$ is smooth and strongly convex and the transition map $\Tr(d;\theta)$ is Lipschitz in both arguments, then repeated risk minimization converges to an equilibrium distribution-classifier pair, coming within distance $\delta$ in $O(\log 1/\delta)$ rounds.
\end{thm}

As defined, these algorithms require direct access to the data distribution.
We also analyze a finite-sample version of RRM and obtain results on the number of datapoints that must be sampled at each round to guarantee linear convergence to the equilibrium pair (see \Thm{finite-sample}).

For some settings, Delayed RRM may require fewer retraining rounds than RRM  until convergence. In \Sec{kGRS}, we compare the two strategies for the scenario of $k$ Groups Respond Slowly (\Ex{kGRS}). This scenario naturally invites a Delayed RRM approach: delay retraining for $k$ rounds until all groups have caught up to the latest classifier. We show that the two algorithms converge to an equilibrium at similar rates in terms of deployment rounds, but if we are only concerned with retraining resources, Delayed RRM is superior.

While converging to an equilibrium pair is a desirable outcome for the institution, this might not be the \textit{optimal} outcome. We formalize the notion of an optimal strategy within our stateful performativity framework. The concept of fixed point distributions is again key to this definition. We show that repeated risk minimization also provides a means to approximate such optimal strategies. 
\begin{thm}[Informal, see Theorem~\ref{thm:rrm_optimality}]
\label{thm:informal_2}
If the loss function $\ell(z;\theta)$ is Lipschitz and strongly convex and the transition map $\Tr(d;\theta)$ is Lipschitz in both arguments, all equilibrium pairs and optimal pairs lie within a small distance of each other. 
\end{thm}

\Thm{informal_1} and \Thm{informal_2}, which we state formally in Section \ref{sec:results}, generalize results of \cite{perdomo2020}~in the stateless framework to our stateful framework. We include complete proofs in the supplementary material.

\subsection{Related Work}

Our work is closely related to that of \cite{perdomo2020}. Various aspects of the stateless performativity framework of \cite{perdomo2020} have been studied, such as stochastic and zeroth-order methods for converging to an equilibrium \parencite{DrusvyatskiyX20, MaheshwariCMSR21, Mendler-DunnerP20}, convergence to the optimal classifier as opposed to an equilibrium  \parencite{IzzoY021, MillerPZ21}, regret minimization \parencite{JagadessanZM22}, and characterization of regions of attraction for different equilibria \parencite{DongR21}. \cite{NarangFDFR22} propose a multi-player performativity framework where the population reacts to competing institutions' actions.  Strategic classification, a term coined by \cite{hardt2016strategic}, is one specific instantiation of performative prediction that has received much attention, e.g., \cite{BechavodLWZ21, ChenLP20, DongR21, HaghtalabILW20, HuIV19, MilliMDH19, Munro20, ShavitEA20, TsirtsisR20, ZrnicMSJ21}. 
Strategic classification studies the behavior of individuals who wish to achieve a more preferable outcome from a classifier by manipulating their attributes without changing their true label. \cite{HuIV19} study the disparate effect of strategic manipulation when groups face different costs to manipulation. In $k$ Groups Respond Slowly (\Ex{kGRS}), we model a different aspect of disadvantage, namely access to information. 

Prior to our work, \cite{LiW21} were the only ones to also consider a stateful setting for performative prediction. 
They study stochastic optimization in the case when the institution samples only one datapoint (or minibatch) at each round, and the updated sample depends both on the current classifier and the previous sample. The samples evolve according to a controlled Markov Chain that depends on the current classifier. The key conceptual difference between the two frameworks is that we update the population-level data based on the previous distribution rather than the realized data from that distribution. Our model subsumes a setting where distributions are updated according to a classifier-dependent Markov chain (see \Ex{markov}).  

\cite{WoodBA21} also propose a more general framework than that of \cite{perdomo2020},~where for every round $t$ a function $\D_t$ maps the classifier to the updated distribution. In contrast to our framework, the updated distributions do not depend on previous distributions. While our framework does not explicitly track time, this can be encoded into the transition function with a simple trick.
The focus of \cite{WoodBA21} is on stochastic optimization, whereas we mainly obtain population-level results.

Following a preliminary version of our work, \cite{RayRDF22} study the geometrically decaying setting of \Ex{GDR} and analyze algorithms that converge to the optimal point. In their work, the institution has oracle access (for a fixed batch size) to either the empirical gradient of the loss or the empirical loss function. They provide high probability finite sample guarantees for both types of oracle access. \cite{IzzoZY21b} study convergence to an optimal point within our framework and provide convergence and sample complexity guarantees, under the assumption that the distributions induced by the classifier are parametric.

Stochastic programming and robust optimization are two general frameworks for modeling optimization problems that involve uncertainty. Within these frameworks, a body of work has studied the case when the system's performance uncertainty depends on the decision variables. Such a setting is usually referred to as decision-dependent distributions or endogenous uncertainty.  We refer to \cite{HellemoBT18} and \cite{LuoM20} and references within for an overview. 

Performative prediction can be seen as a special case of reinforcement learning, but the former aims to abstract different phenomena. Performative prediction is designed to capture settings where supervised learning with retraining is a common approach. We explore when natural extensions of supervised techniques to handle performativity (namely RRM) can still succeed and the full strength and complexity of reinforcement learning techniques (such as learning the Q-value function) are unnecessary. For instance, RRM is an algorithm that explicitly only ``exploits'' and never ``explores.''

\section{FRAMEWORK AND MAIN RESULTS}\label{sec:results}

In this section, we formally state our main results and the relevant definitions. 
We parameterize machine learning models by real-valued vectors $\theta \in \Theta$. In round $t$, the institution chooses a classifier $\theta_t$.  The adversary responds with a distribution $d_t \in \Delta(\mc{Z})$ over instances $z = (x,y) \in \mathbb{R}^{m-1} \times \mathbb{R}$ of feature-label pairs. The adversary is restricted to pick its distribution according to a deterministic transition map:
\[
\Tr: \Theta \times \Delta(\mc{Z}) \rightarrow \Delta(\mc{Z}),
\]
so that $\Tr(d_{t-1}, \theta_t)  = d_t$. 
We assume that an initial distribution $d_0$ is publicly known.

\begin{algorithm}
\caption{Performative prediction with state}
\begin{algorithmic}[1]
    \State Initial distribution $d_0\in\Delta(\mc{Z})$  \Comment{Publicly known}
    \For{t = 1,2,\ldots}
        \State Institution publishes $\theta_t\in\Theta$.
        \State Adversary computes $d_t=\Tr(d_{t-1};\theta_t)$.
        \State Institution observes $d_t$, suffers loss $\E_{z\sim d_t}[\ell(z;\theta_t)]$.
    \EndFor
\end{algorithmic}
\end{algorithm}

\subsection{Repeated Risk Minimization and Stable Pairs}

\cite{perdomo2020}~propose the following strategy for the institution: at each round, play the classifier that minimizes loss on the previous distribution. We investigate the same strategy in our stateful framework. 

\begin{defn}[Repeated Risk Minimization (RRM)] \label{defn:rrm}
	Denote by $G(d)$ the risk minimizer\footnote{For the scenarios we consider, the set of minimizers will be non-empty. When the set has more than one element, RRM chooses a value from the set arbitrarily.}:
	\begin{equation*}
	G(d) := \underset{\theta'}{\mathrm{argmin}} \: \underset{z \sim d}{\E}  \ell(z; \theta').
	\end{equation*}
	Following the notation of Algorithm 1, at round $t$, the institution updates its classifier to $\theta_{t} = G(d_{t-1})$. 
\end{defn}

For clarity, we define additional notation wrapping the institution's and adversary's actions into one step.

\begin{defn}[RRM map]
    Define the RRM map $f \colon \Delta(\mathcal{Z}) \times \Theta \rightarrow \Delta(\mathcal{Z}) \times \Theta$ as:
\begin{equation*}
    \label{f_map}
    f(d, \theta) = (\Tr(d, \theta), G(\Tr(d, \theta))). 
\end{equation*}
In our game, $f(d_{t-1}, \theta_t) = (d_t, G(d_t)) = (d_t, \theta_{t+1})$.
\end{defn}

We consider two sufficient conditions for the convergence of RRM in objective value: approaching a \emph{fixed point distribution} and approaching a classifier that is optimal for this distribution. %

\begin{defn}[Fixed point distribution]\label{def:fixed_point}
A distribution $d_\theta$ is a \emph{fixed point for $\theta$} if $\Tr(d_\theta, \theta) = d_\theta$.
\end{defn}

\begin{defn}[Stable Pair]
	 A distribution-classifier pair $(d_{\mathrm{S}}, \theta_\mathrm{S})$ is a \emph{stable pair} if the following hold:
	 \begin{enumerate}
	     \item %
	     $d_\ST$ is a fixed point distribution for $\theta_\ST$. 
	     \item $\theta_\mathrm{S} = G(d_\mathrm{S})$, i.e. $\theta_{\mathrm{S}}$ minimizes the loss on $d_{\mathrm{S}}$.
	 \end{enumerate}
\end{defn}

Once the game approaches the distribution $d_\ST$, the institution can repeatedly play $\theta_\ST$ with no need for retraining, while incurring the lowest possible loss on the distribution $d_\ST$. It is not obvious, however, that such stable pairs exist for every setting. Nevertheless, we shows sufficient conditions on the loss and transition function for RRM to converge to a stable pair.

\begin{defn}[Strong convexity] A loss function $\ell(z;\theta)$ is $\gamma$-strongly convex if, for all $\theta, \theta' \in \Theta$ and $z \in \mc{Z}$,
	\begin{equation*}
	\ell(z;\theta) \geq \ell(z;\theta') + \nabla_\theta \ell(z;\theta')^\top(\theta - \theta') + \frac{\gamma}{2} \lVert\theta - \theta'\rVert_2^2.
	\end{equation*}
\end{defn}

\begin{defn}[Smoothness] 
	
	A loss function $\ell(z;\theta)$ is $\beta$-jointly smooth if the gradient with respect to $\theta$ is $\beta$-Lipschitz in $\theta$ and $z$, i.e.,
	\begin{align*}
	\lVert \nabla_\theta \ell(z;\theta) - \nabla_\theta \ell(z;\theta') \rVert_2 \leq \beta \lVert \theta - \theta' \rVert_2,  \\
	\lVert \nabla_\theta \ell(z;\theta) - \nabla_\theta \ell(z';\theta) \rVert_2 \leq \beta \lVert z - z'\rVert_2, 
	\end{align*}
	for all $\theta, \theta' \in \Theta$ and $z, z' \in \mc{Z}$. 
\end{defn}

The next example shows that, without the right interplay of the above parameters, there are settings for which RRM may not converge.

\begin{ex}[RRM may not converge]\label{ex:rrm_may_not_converge}
\emph{Take the loss function to be the squared loss $\ell(z; \theta) = (y - \theta)^2$ for $\theta \in [1, \infty)$. The loss function is $\beta$-jointly smooth and $\gamma$-strongly convex, with $\beta = \gamma = 2$. Consider the transition map $\Tr(d; \theta) = 1 + \eps d + \eps \theta$, which operates on point mass distributions $d \in [1, \infty)$ of the outcome $y$. Clearly, the transition function $\Tr$ is $\eps$-jointly sensitive. Finally, let some $d_0 \in [1, \infty)$ be the starting distribution of the game.} 

\emph{When the institution uses RRM starting from $d_0$, we get that:}
\begin{align*}
    \theta_{t+1} &= G(d_t) = \underset{\theta}{\mathrm{argmin}} \: \underset{z \sim d_t}{\E} \ell(z; \theta) = d_t,\\
    d_{t+1} &= \Tr(d_t; \theta_{t+1}) = 1 + \eps d_t + \eps \theta_{t+1}.
\end{align*}
\emph{Hence,~$\theta_{t+2} = d_{t+1} = 1 + \eps d_t + \eps \theta_{t+1} = 1 + 2\eps \theta_{t+1}.$}

\emph{The distance between two successive classifiers is $|\theta_{t+2} - \theta_{t+1}| = |(1+2\eps\theta_{t+1}) - (1+2\eps\theta_{t})| = 2\eps|\theta_{t+1} - \theta_{t}|$.} 

\emph{If we only require $\eps < \frac{\gamma}{\beta} = 1$, then whenever $\eps > \frac{1}{2}$, the sequence of $\theta$'s produced by RRM fails to converge. When $\eps < \frac{1}{1+\beta/\gamma} = \frac{1}{2}$, 
the sequence converges. $\square$}
\end{ex}

Our main result is sufficient conditions for the convergence of RRM in the stateful framework. Endow the space $ \Delta(\mathcal{Z}) \times \Theta$ with the product metric $\mathrm{dist}(\cdot,\cdot)$, so that
\begin{equation*}
    \mathrm{dist}((d, \theta), (d', \theta')) = \mc{W}_1(d, d') + \lVert \theta-\theta'\rVert_2.
\end{equation*}

\begin{thm}
	\label{thm:iterative_risk_min}
	Suppose the transition map $\Tr(;)$ is $\eps$-jointly sensitive and the loss function $\ell(z; \theta)$ is $\beta$-jointly smooth and $\gamma$-strongly convex. Let~$\alpha = \eps(1+\frac{\beta}{\gamma})$. Then for the RRM map $f$, and all $d, d' \in \mc{Z}$ and $\theta, \theta' \in \Theta$, it holds that
	\begin{enumerate}[(a)]
		\item  $\mathrm{dist}(f(d, \theta), f(d', \theta')) \leq \alpha \cdot \mathrm{dist}((d, \theta), (d', \theta'))$.
		\item  In particular, if $\alpha < 1$, then $f$ has a unique fixed point which is a stable pair with respect to $\Tr(;)$. RRM will converge to this stable pair at a linear rate. 
	\end{enumerate}
\end{thm}

In the introduction, we discussed Delayed RRM, which is a more complex approach to convergence in the stateful setting.
Since RRM converges faster, we keep further discussion of Delayed RRM for the Supplementary Material, where we state and prove \Thm{delayed_rrm}.

Finally, we analyze the empirical counterpart of RRM, where at each round the institution only has access to a finite sample from the distribution. 
\begin{defn}[Repeated Emprical Risk Minimization (RERM)]
At timestep $t$, sample $n_t$ samples from $d_{t-1}$. Let $\widetilde{d}_{t-1}$ be the uniform distribution on the $n_t$ samples from $d_{t-1}$. Update the classifier to $\theta_t = G( \widetilde{d}_{t-1})$. 
\end{defn}

\begin{thm} \label{thm:finite-sample}
Suppose that the loss $\ell(z, \theta)$ is $\beta$-jointly Lipschitz and $\gamma$-strongly convex and there exist $\alpha > 1, \mu > 0$ such that $\int_{\mathbb{R}^m} e^{\mu |x|^\alpha} d'(dx)$ is finite for all $d' \in \Delta(\mc{Z})$.  Fix $\delta \in (0, 1)$ to be a radius of convergence. Consider running RERM with $n_t = O\Bigl( \frac{\log(t/p) }{(\eps(1+\frac{\gamma}{\beta})\delta)^m} \Bigr)$ samples at time $t$. If the transition map $\Tr(;)$ is $\eps$-jointly sensitive and $2\eps\Bigl( 1 + \frac{\beta}{\gamma}\Bigr) < 1$, then with probability $1-p$, the iterates of RERM are within a radius $\delta$ of a stable pair  for $t \geq  \Bigl(1 - 2\eps\Bigl(1+\frac{\beta}{\gamma}\Bigr) \Bigr) O( \log(1/\delta))$. 

\end{thm}

Our conditions for convergence of RRM and RERM are similar to, but stricter than, those of the setting of \cite{perdomo2020}. In particular, their results for convergence of RRM only require that $\eps < \frac{\gamma}{\beta}$. \Ex{rrm_may_not_converge} shows that our requirement is necessary. 

\subsection{Optimality}

\Thm{iterative_risk_min} guarantees that RRM converges to an equilibrium,
but this stable pair might not be optimal with respect to loss. 
In fact, it is not obvious how to define the notion of ``optimal classifier,'' since the sequence of distributions played by the adversary depends on the initial distribution.
Therefore, to define optimality, we restrict our attention to scenarios where the transition map $\Tr(;)$ is $\eps$-jointly sensitive with $\eps < 1$. 
In this setting, repeatedly applying the same classifier $\theta$ causes convergence to a single distribution (\Def{fixed_point}).

\begin{claim}
\label{clm:banach}
If the transition map $\Tr(;)$ is $\eps$-jointly sensitive with $\eps < 1$, then for each $\theta \in \Theta$, there exists a unique fixed point distribution $d_\theta$. For any starting distribution $d_0$, iterated application of the same classifier $\theta$ will result in a sequence of distributions that converges to $d_\theta$ at a linear rate.
\end{claim}

\Clm{banach} follows immediately from Banach's fixed point theorem. 

Our definition of the optimal strategy considers the ``long-run'' loss of a fixed classifier. Assume the institution plays the same fixed classifier $\theta$ for all rounds of the game. We measure the long-run loss of $\theta$ as the loss on its corresponding fixed point distribution $d_\theta$. The optimal $\theta$ is the one which minimizes its long-run loss. 

\begin{defn}[Optimality] The long-run loss of a classifier $\theta$ is the loss $\underset{z \sim d_{\theta}}{\E} \ell(z; \theta)$, where $d_\theta$ denotes the unique fixed point distribution for the classifier $\theta$.  A classifier $\theta_{\mathrm{OPT}}$ is optimal if it achieves the minimum long-run loss amongst all classifiers in $\Theta$. 
\end{defn}

If an institution had prior knowledge of the transition map, a reasonable strategy
would be to play the fixed classifier $\theta_\OPT$ for all rounds of classification. We note that if $\Tr(;)$ is $\eps$-jointly sensitive with $\eps \ge 1$ then  $\theta_{\mathrm{OPT}}$ may not be defined.

Our definitions of stability and optimality generalize those of \cite{perdomo2020}~for the stateless framework. As pointed out in \cite{perdomo2020}, for a given setting, the optimal classifier does not necessarily coincide with a stable classifier. Our next result shows that RRM approximately approaches optimal classifiers. 

\begin{thm}
    \label{thm:rrm_optimality}
    Suppose that the loss $\ell(z;\theta)$ is $L_z$-Lipschitz, $\gamma$-strongly convex, and that the transition map is $\eps$-jointly sensitive with $\eps < 1$. Then for every stable classifier $\theta_{\mathrm{S}}$ and optimal classifier $\theta_{\mathrm{OPT}}$ it holds
    \begin{equation*}
        \lVert \theta_{\mathrm{OPT}} - \theta_{\mathrm{S}} \rVert_2 \leq \frac{2L_z\eps}{\gamma(1-\eps)}.
    \end{equation*}
\end{thm}

\begin{figure*}[h!]
\vspace{.3in}
\centering
\includegraphics[width=\linewidth]{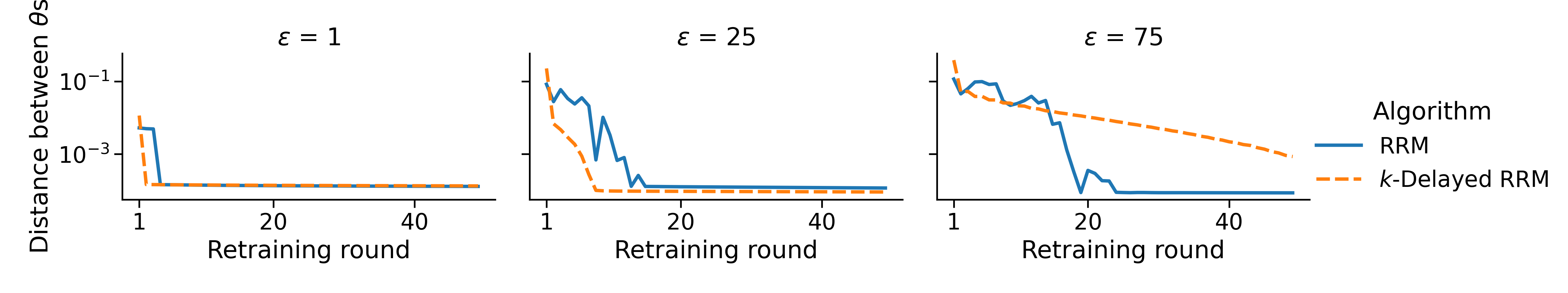} 
\vspace{.3in}
\caption{Convergence of RRM and $k$-Delayed RRM for varying values of $\eps$ and $k=3$. The horizontal axis shows the number of \emph{retraining rounds} and the vertical axis shows the distance between successive $\theta$'s.}
\label{fig:convergence}
\end{figure*}

\section{$k$ GROUPS RESPOND SLOWLY}\label{sec:kGRS}

Consider the setting of \Ex{kGRS}, where the target population contains $k$ distinct subpopulations.
The $j$-th subpopulation, for $j \in [k]$, responds strategically to the classifier from $j$ rounds ago.
This provides a simple model for investigating performativity in settings where information propagates at different rates.
For distribution $d$, let $d^{|j}$ be the distribution conditioned on being in group $j$, and denote the mixture by $d = (d^{|1}, d^{|2}, \dots, d^{|k})$. 
Then the transition function is
\[
\Tr((d^{|1}_t, \dots, d^{|k}_t), \theta_t) = (\D(\theta_t), d^{|1}_t, \dots, d^{|k-1}_{t}).
\]

We compare two algorithms in this setting: RRM and $k$-Delayed RRM (\Alg{kDelayed}), which in this setting is similar to Delayed RRM.
This algorithm updates the classifier $\theta$ only every $k$ rounds, after all groups have caught up to the latest deployed classifier and thus have the same distribution.  

\begin{algorithm}
\caption{$k$-Delayed RRM}
\begin{algorithmic}[1]
    \State \textbf{input:} number of groups $k \in \mathbb{N}$, initial distribution $d_0\in\Delta(\mc{Z})$ 
    \State Let $\theta = G(d_0)$; publish $\theta$. 
    \For{t = 1,2,\ldots}
        \State Observe $d_t$. \Comment{ $d_t= \Tr(d_{t-1};\theta)$}.
        \State If $t \mod k = 0$, update $\theta = G(d_t)$. 
        \State Publish $\theta$. 
    \EndFor
\end{algorithmic}
\label{alg:kDelayed}
\end{algorithm}

Effectively, \Alg{kDelayed} performs RRM on the map $\D(\theta)$ every $k$ rounds, which is the setting of \cite{perdomo2020}. Thus, its rate of convergence is $k$ times of the rate of convergence of RRM in the stateless framework where $k=1$. 

\Thm{kGRS} states the performance of the two algorithms.
We note that the transition map defined above is not $\eps$-jointly sensitive for any $\eps<1$, so our proof of convergence of RRM requires additional analysis beyond \Thm{iterative_risk_min}.
Furthermore, note that the assumption on $\D(\theta)$ corresponds to the notion of \emph{sensitivity} of \cite{perdomo2020}.

\begin{thm}\label{thm:kGRS}
Suppose that, for some $\eps>0$ and for all $\theta, \theta'\in \Theta$, the deterministic strategic response map satisfies $\W_1(\D(\theta),\D(\theta'))\le \eps \lVert \theta - \theta'\rVert_2$.
Additionally suppose that the loss function $\ell(z;\theta)$ is $\beta$-jointly smooth and $\gamma$-strongly convex. For $\eps < \frac{\gamma}{\beta}$, the iterates of $k$-Delayed RRM and RRM converge to $\theta_\ST$ at rate $k(1-\eps \frac{\beta}{\gamma})^{t}$.
\end{thm}

One advantage of \Alg{kDelayed} is that it can require fewer retraining rounds, possibly requiring fewer computational resources. 
On the other hand, RRM does not require prior knowledge of the number of groups $k$.

\section{SIMULATION}\label{sec:simulation}

Strategic classification studies the behavior of individuals who wish to achieve a more preferable outcome from a classifier by manipulating their attributes without changing their true label \parencite{hardt2016strategic}. It is one instantiation of performative prediction. We adapt a simulation of loan applications in \cite{perdomo2020}, implemented using the \texttt{WhyNot} Python package \parencite{miller2020whynot}, and enrich it with state. We demonstrate the convergence of RRM and $k$-Delayed RRM for the scenario of $k$ Groups Respond Slowly (\Ex{kGRS})
in a credit score setting.\footnote{\url{https://github.com/shlomihod/performative-prediction-stateful-world}} We run our experiments on a desktop computer.

The distribution for each of the $k$ groups is deterministically initialized as an instance of the baseline distribution.
The baseline distribution is the uniform distribution over samples in Kaggle's \emph{GiveMeSomeCredit} dataset \parencite{givemesomecredit} consisting of $N=18,357$ individuals. Hence, there are $kN$ individuals in the whole population.

An individual's strategic response is based on cost and utility functions that take into account the published classifier and the attributes of the individual taken from the \emph{GiveMeSomeCredit} dataset. However, the groups may respond to different $\theta$'s at the same round.

The parameter $\eps$ controls the strength of the strategic response; larger values allow greater manipulation. We run the simulation for $k$ Groups Respond Slowly using both RRM and $k$-Delayed RRM, for 
$k=3$, $\varepsilon \in \{1, 25, 75\}$, and 300 rounds. Note that for $k=1$ the two algorithms are identical and the simulation is the same as the stateless setting of \cite{perdomo2020}.
Refer to Appendix \ref{app:simulation} for a detailed description of the simulation. 

\Fig{convergence} shows the game dynamics when the institution uses the two different algorithms. 
The vertical axis shows the distance between successively trained classifiers. Recall that in $k$-Delayed RRM the classifier is retrained once in $k$ rounds (in constrat to RRM where retraining happens every round). Therefore, to depict convergence of the algorithms we plot the distance between successively trained classifiers against the number of \emph{retraining} rounds.

Since the initial distribution consists of $k$ copies of the \emph{GiveMeSomeCredit} dataset, the \emph{same} classifier is deployed at the first retraining round for each $\eps$, and only the most advantaged group responds strategically. For larger $\eps$, the individuals from the most advantaged group respond with a larger update to their features, and therefore in the second round, the institution trains a classifier which is further from the first. 
In Figure~1 we can see a lower value of $\norm{\theta_{2} - \theta_1}$ for $\varepsilon=1$ than for the settings of $\varepsilon=25$ and $\varepsilon=75$.

For $\eps \in \{1, 25\}$, we see that $k$-Delayed RRM converges faster than RRM in terms of number of retraining rounds. Although not depicted here, RRM was superior to $k$-Delayed RRM in terms of number of rounds (elapsed time) until convergence for all parameter settings we considered. Thus, an important consideration for practitioners in choosing an algorithm is whether elapsed time or retraining resources used until convergence is the more valuable metric.

\begin{figure*}[th!]
\centering
        \includegraphics[width=\textwidth]{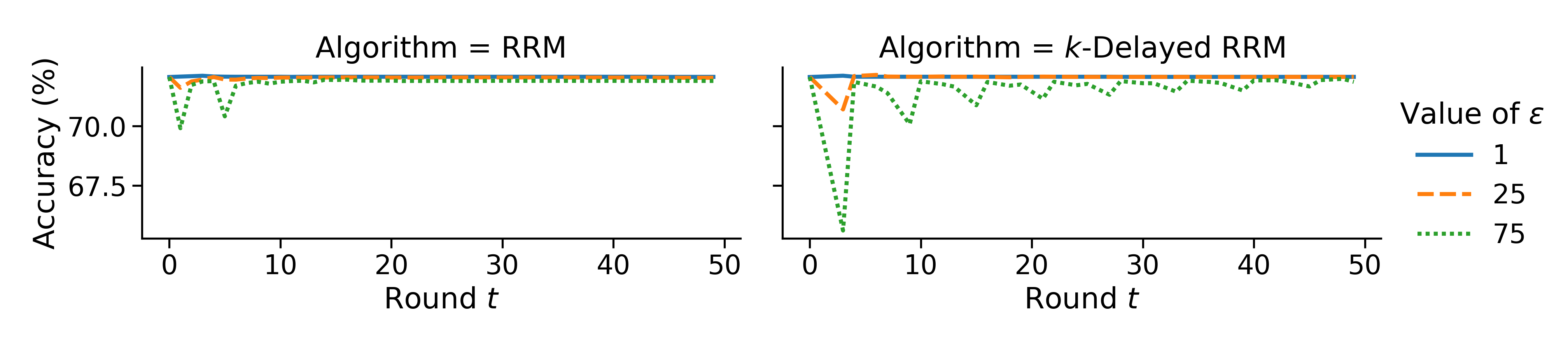}
         \vspace{.05in}
         \caption{Accuracy of RRM and $k$-Delayed RRM for different $\eps$ and $k=3$. The horizontal axis shows the number of \emph{rounds} and the vertical axis shows the accuracy of the published model after the strategic response.}
         \label{fig:utility}
\end{figure*}

Interestingly,
as demonstrated in \Fig{utility}, for $\eps<75$
both algorithms reach high accuracy much faster than they converge to the stable pair, and there is little difference 
between the accuracy dynamics of RRM and $k$-Delayed RRM.
For $\varepsilon = 75$, the accuracy increase of $k$-Delayed RRM is much slower than that of RRM. In \Fig{utility}, at round $t=0$ we show the accuracy of the first deployed classifier \emph{before} the strategic response, i.e. the accuracy with respect to the baseline distribution. In all other rounds, the accuracy of the classifier is shown \emph{after} the strategic response.

\begin{figure*}[th!]
         \centering
         \includegraphics[width=\textwidth]{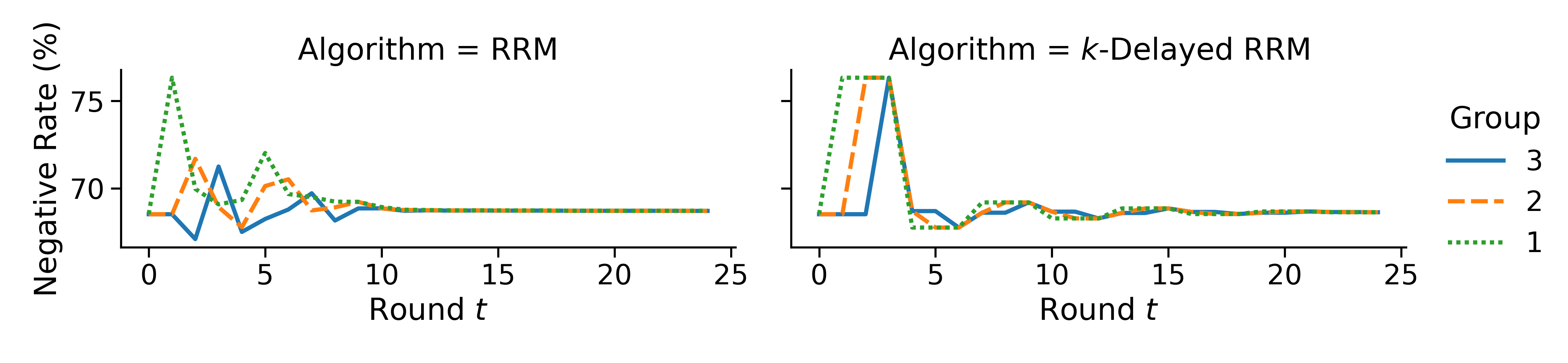}
         \vspace{.05in}
         \caption{Negative rate of RRM and $k$-Delayed RRM for $\varepsilon = 25$ and $k=3$. The horizontal axis shows the number of \emph{rounds} and the vertical axis shows the negative rate of the published model after the strategic response.}
         \label{fig:fairness}
\end{figure*}

Finally, we study the game dynamics from the perspective of the target population, stratified by the $k$ groups.
Due to the different rate of information acquisition, there is a hierarchy of advantage among the groups. Intuitively, the group  that responds first to the latest deployed classifier has an \emph{advantage}\footnote{We stress that the operationalization of disadvantage and access to information is deliberately abstracted in our model, and we do not attempt to fully capture these complex societal concepts.} compared to other groups. We investigate if this hierarchy of advantage translates into a disparity between groups in the benefit they achieve from their strategic response. This raises a question of whether the choice of algorithm can mitigate the disparity effect. 
We conjecture  that since $k$-Delayed RRM allows the groups to reach a homogeneous distribution, it is preferable to RRM in terms of per-group disadvantage.  

We investigate the per-group \emph{negative rate} (NR),  the proportion of individuals that are predicted unlikely to default, as a measure of the groups' benefit from their strategic response. We focus on $\varepsilon = 25$ in our analysis. \Fig{fairness} shows that in the initial rounds ($1 \leq t < 10$), before NR converges to the same value for all groups, there is a noticeable difference between the groups for both algorithms. When accumulated over the first 10 rounds, the most advantaged group has $2.37\%$ and $2.09\%$ more negative predictions than the most disadvantaged group for RRM and $k$-Delayed RRM, respectively. It appears that in this simulation there is no clear advantage to either one of the algorithms in terms of per-group disparity.

The definition of $k$ Groups Respond Slowly allows for different initial distribution for each group. In Appendix~\ref{app:sim-init}, we run additional experiments with varying initial distributions for the groups to supplement our existing results.

\section{DISCUSSION}\label{sec:conclusion}

This work models the important role that the history of  institutional predictions plays in shaping the behavior of individuals. 
The addition of state to the formal model of performativity opens up a new venue for discussing the social impact of machine predictions. Examples include the structural changes that enable individuals to succeed under such modes of classification and the disparate impact of predictions on groups over time. We believe the setting of \Ex{kGRS}, $k$ Groups Respond Slowly, provides an excellent starting point for future investigation of the interaction between performativity and fairness in the presence of information disparities.

The theoretical assumptions in this work constrain either the loss function or the transition function.
While an institution controls the loss function, there is no clear way to verify in practice the 
existence of a fixed, deterministic transition function, let alone its sensitivity.
Our theoretical guarantees are thus best interpreted in practice as conditional statements: if the world's response is not too sensitive to prior history or the institution's choices, the simple and widely used heuristic of RRM is sensible.

This work focuses on the goal of convergence, but if convergence is not a priority for the institution then it may be interesting to study which measures of the institution's success best apply to the setting of performativity. Regret is widely studied in online learning, but lacks a clear interpretation in settings where the adversary can adapt to the player.  Empirically, studying other applications where performativity arises could provide important theoretical insight into this phenomenon.

\section{SOCIETAL IMPACT}\label{sec:impact}

Our work assumes a simple and abstract model of repeated decision-making.
While the study of performativity in prediction may have wide social effects in general, we do not believe this paper will have direct ethical or social consequences.

\subsubsection*{Acknowledgements}
The authors would like to thank Adam D. Smith for fruitful conversations and guidance, and Ran Canetti for helpful comments. 

\printbibliography

\clearpage
\appendix

\thispagestyle{empty}

\onecolumn \makesupplementtitle

\section{PROOFS OF THEOREMS}\label{sec:proofs}

We first state some key lemmas used in the proofs of our theorems. The second lemma follows directly from the proof of Theorem 3.5 of \cite{perdomo2020}.

\begin{lemma}[\cite{bubeck15}]
\label{lem:bubeck}
If $g$ is convex and $\Omega$ is a closed convex set on which $g$ is differentiable, and 
	\begin{equation*}
	x_* \in \underset{x\in \Omega}{\mathrm{argmin}} \: g(x),
	\end{equation*} 
	then	$(y - x_*)^\top \nabla g(x_*) \geq 0$ for all $y \in \Omega$.

\end{lemma}

\begin{lemma}[\cite{perdomo2020}]\label{lem:perdomo}
	\label{lem:perdomo_main_theorem}
	Suppose the loss function $\ell(z;\theta)$ is $\beta$-jointly smooth and $\gamma$-strongly-convex. For any two distributions $d, d' \in \Delta(\mc{Z})$, risk minimization satisfies
	\begin{equation*}
		\lVert G(d) - G(d') \rVert_2 \leq \frac{\beta}{\gamma}\mc{W}_1 (d, d'). 
	\end{equation*}
\end{lemma}

\begin{lemma}
\label{lem:sensitivity}
Suppose the map $\Tr(;)$ is $\eps$-jointly sensitive with $\eps < 1$. Then for any $\theta_1, \theta_2 \in \Theta$ and their corresponding fixed point distributions $d_1, d_2$ it holds that
\begin{equation*}
    \mc{W}_1(d_1, d_2) \leq \frac{\eps}{1-\eps} \lVert \theta_1 - \theta_2 \rVert_2.
\end{equation*}
\end{lemma}
\begin{proof}
By the definition of fixed point distributions,  $\mc{W}_1(d_1, d_2) = \mc{W}_1(\Tr(d_1, \theta_1), \Tr(d_2, \theta_2))$. Since the  transition map is $\eps$-jointly sensitive, we obtain
\begin{equation*}
    \mc{W}_1(d_1, d_2) \leq \eps \mc{W}_1(d_1, d_2) + \eps \lVert \theta_1 - \theta_2 \rVert_2.
\end{equation*}
The lemma follows from the equation above. 
\end{proof}

\subsection{Proof of \Thm{delayed_rrm_informal}}

In this section, we formally state our Delayed RRM algorithm (\Alg{delayed}) and prove \Thm{delayed_rrm_informal}, restated formally below in \Thm{delayed_rrm}. 

\begin{algorithm}[H]
	\caption{Main procedure \textbf{Delayed RRM}}
	\begin{algorithmic}[1]
		\Require{radius $\delta$, loss function $\ell$, initial distribution $d_0$, sensitivity $\eps \in (0,1)$}
		\State Let $ d \leftarrow d_0$. 
		\State Initialize $\theta \leftarrow \mathbf{0}$. 
		\Loop $\:t$ times:
		\State Calculate $\theta = \underset{\theta'}{\arg\min} \underset{z \sim d}{\E}  \ell(z;\theta')$. \label{step:calc_theta}
		\State Update $d \leftarrow \mathbf{RepeatedDeployment}(d, \theta, \delta, \eps)$.
		\EndLoop
	\end{algorithmic}
	\label{alg:delayed}
\end{algorithm}

\begin{algorithm}[H]
	\caption{Sub-procedure $\mathbf{RepeatedDeployment}$}
	\begin{algorithmic}[1]
		\Require{initial distribution $d_0$, classifier $\theta$, radius $\delta$, sensitivity $\eps \in (0,1)$}.

		\State Publish $\theta$. 
		\State Observe $d_1 \leftarrow \Tr(d_0; \theta)$. 
		\State Let $d_{\mathrm{current}} \leftarrow d_1$. 
		\State Let $r = \log^{-1}(\frac{1}{\eps})\log(\frac{\mc{W}_1(d_0, d_1)}{\delta})$. 
		\Loop $\: r$ times:
		\State Publish $\theta$.
		\State Observe $d_{\mathrm{current}} \leftarrow \Tr(d_{\mathrm{current}}; \theta)$.
		\EndLoop
		\State Return $d_{\mathrm{current}}$. 
	\end{algorithmic}
\label{alg:d_approx_procedure}
\end{algorithm}

\begin{thm}
	\label{thm:delayed_rrm}
	Suppose the loss function $\ell(z;\theta)$ is $\beta$-jointly smooth and $\gamma$-strongly convex. Let $\theta_1, \dots, \theta_t$ denote the classifiers obtained in each iteration of Step \ref{step:calc_theta} of \Alg{delayed}.  If the transition map $\Tr(\cdot; \cdot)$ is $\eps$-jointly sensitive with $\eps < 1$ and $\frac{\eps}{1-\eps} < \frac{\gamma}{2\beta}$, then:
	
	\begin{enumerate}[(a)]
		\item $\lVert \theta_t - \theta_\ST \rVert_2 \leq \delta$ for $t \geq \bigl( 1 - \frac{2\eps\beta}{\gamma(1-\eps)}  \bigr)^{-1} \log(\frac{\lVert \theta_0 - \theta_\ST \rVert_2}{\delta})$.
		
		\item Each iteration of the main $\mathbf{loop}$ in  \Alg{delayed} consists of $r \leq \log(\frac{1}{\eps})^{-1}\log(\frac{\mathrm{diam}(\Delta(\mc{Z}))}{\delta})$ deployments of the same classifier, where $\mathrm{diam}(\Delta(\mc{Z}))$ denotes the largest $\mc{W}_1$ distance between two distributions in $\Delta(\mc{Z})$. Thus, $O(\log^2(\frac{1}{\delta}))$ deployments are needed for $\theta_t$ to be within distance $\delta$ of $\theta_\ST$. 
	\end{enumerate}
\end{thm}

We first prove an auxiliary lemma regarding the \textbf{RepeatedDeployment} procedure.

\begin{lemma}
	\label{lem:d_approx_radius}
	Given a classifier $\theta$, denote by $\widetilde{d}_\theta$ the distribution returned from $\mathbf{RepeatedDeployement}(d_0, \theta, \delta, \eps)$ for any $d_0 \in \Delta(\mc{Z})$. Let $d_\theta$ be the fixed point distribution for $\theta$. If $\eps <1$, then
	\begin{equation*}
	\mc{W}_1(\widetilde{d}_\theta, d_\theta) \leq \frac{\eps}{1-\eps}\delta.
	\end{equation*}
	
	\begin{proof}
		For a fixed $\theta$ and $\eps<1$, the map $\Tr(\cdot;\theta)$ is contractive with Lipschitz coefficient $\eps$ and has a unique fixed point $d_\theta$.  Note that $\widetilde{d}_\theta = \Tr^{r+1}(d_0; \theta)$, where the transition map is applied $r+1$ times with the same classifier $\theta$.  Let $d_1 =\Tr(d_0; \theta)$. It is easy to see that
		\begin{equation}
		\label{eq:distance_to_d_infty}
			\mc{W}_1(\widetilde{d}_\theta, d_\theta) \leq \frac{\eps^{r+1}}{1-\eps}\mc{W}_1(d_0, d_1).
		\end{equation}
		For $r  = \log^{-1}(\frac{1}{\eps})\log(\frac{\mc{W}_1(d_0, d_1)}{\delta})$, we obtain $\eps^{r} =  \frac{\delta}{\mc{W}_1(d_0, d_1)}$.
		Plugging the value of $\eps^r$ into \eqref{eq:distance_to_d_infty} concludes the proof.
	\end{proof}
\end{lemma}

\begin{proof}[Proof of \Thm{delayed_rrm}]
	We prove part (a). As before, let $\widetilde{d}_\theta$ denote the distribution returned from a single call of $\mathbf{RepeatedDeployement}(\cdot, \theta, \delta, \eps)$. Note that
		\begin{equation*}
		\theta_{i+1} = G(\widetilde{d}_{\theta_i}).
		\end{equation*}
	Recall that $\theta_{\mathrm{S}} = G(d_{\ST})$. Then, \Lem{perdomo} gives
	\begin{equation}
	\label{eq:bound_g_by_w}
	\norm{\theta_{i+1} - \theta_\ST} = 	\lVert G(\widetilde{d}_{\theta_i}) -  G(d_{PS})\rVert_2 \leq \frac{\beta}{\gamma}\cdot\mc{W}_1(\widetilde{d}_{\theta_i},d_\ST ).
	\end{equation}
	We bound the distance $\mc{W}_1(\widetilde{d}_{\theta_i},d_\ST )$. Let $d_{\theta_i}$ be the fixed point distribution for $\theta_i$. By the triangle inequality,
	\begin{equation}
	\label{eq:triangle_ineq}
		\mc{W}_1(\widetilde{d}_{\theta_i},d_\ST ) \leq 
		\mc{W}_1(\widetilde{d}_{\theta_i}, d_{\theta_i}) + 
		\mc{W}_1(d_{\theta_i}, d_{\ST}).  
	\end{equation}
	The first term of the sum in \eqref{eq:triangle_ineq} can be bound using \Lem{d_approx_radius}, whereas the second term can be bounded using \Lem{sensitivity} and the fact that $d_{\theta_i}$ and $d_\ST$ are fixed point distributions. We obtain
	\begin{equation}
	\label{eq:triangle_ineq_2}
		\mc{W}_1(\widetilde{d}_{\theta_i},d_\ST ) \leq 
		\frac{\eps}{1-\eps}\delta + \frac{\eps}{1-\eps}\lVert \theta_{i} - \theta_{\mathrm{S}} \rVert_2. 
	\end{equation}
	
	We consider two cases. In the case when $\lVert \theta_i - \theta_{\mathrm{S}} \rVert_2> \delta$, we show that in the next retraining round, the classifier $\theta_{i+1}$ moves closer to $\theta_{\mathrm{S}}$. Replace $\delta < \lVert \theta_i - \theta_{\mathrm{S}}\rVert_2$ in \eqref{eq:triangle_ineq_2}. Then from \eqref{eq:bound_g_by_w} we obtain  $\lVert \theta_{i+1} - \theta_{\mathrm{S}} \rVert_2 \leq \frac{2\eps}{1-\eps}\frac{\beta}{\gamma} \lVert \theta_i - \theta_{\mathrm{S}} \rVert_2$, i.e., contraction happens in iteration $i+1$. In the case when  $\lVert \theta_i - \theta_{\mathrm{S}} \rVert_2 \leq \delta$, combining \eqref{eq:bound_g_by_w}  and \eqref{eq:triangle_ineq_2} yields
	\begin{equation*}
		\lVert \theta_{i+1} - \theta_{\mathrm{S}} \rVert_2 \leq \frac{2\eps}{1-\eps}\frac{\beta}{\gamma}\delta \leq \delta. 
	\end{equation*} 
	This shows that the classifier $\theta_{i+1}$ does not leave the ball of radius $\delta$ around $\theta_{\mathrm{S}}$. The two cases combined give that for $t \geq \bigl( 1 - \frac{2\eps\beta}{\gamma(1-\eps)}  \bigr)^{-1} \log(\frac{\lVert \theta_0 - \theta_{\mathrm{S}} \rVert_2}{\delta})$ iterations we have
	\begin{equation*}
		\lVert \theta_t - \theta_{PS} \rVert_2 \leq \biggl(   \frac{2\eps\beta}{\gamma(1-\eps)} \biggr)^t \lVert \theta_0 - \theta_{\mathrm{S}}\rVert_2 \leq \delta,
	\end{equation*}
	which concludes the proof of part (a). Part (b) is clear from the statement of the subprocedure $\mathbf{RepeatedDeployement}$.
\end{proof}

\subsection{Proof of \Thm{iterative_risk_min}}

	\begin{proof}[Proof of \Thm{iterative_risk_min}]
	    Note that if part (a) holds, then part (b) follows from the fact that the map $f$ is contractive with contraction coefficient $\eps(1+\frac{\beta}{\gamma}) < 1$. By the Banach fixed point theorem, $f$ has a unique fixed point. Suppose $(d^*, \theta^*)$ is the fixed point of $f$, so that $f(d^*, \theta^*) = (d^*, \theta^*)$. This point is also a stable pair for it satisfies $d^* = \Tr(d^*, \theta^*)$ and $\theta^* = G(\Tr(d^*, \theta^*)) = G(d^*)$. 
	    
	    We now show part (a). We simplify notation and let $G(d, \theta) := G(\Tr(d, \theta))$. By definition of $f$, it holds
	    \begin{align*}
	        \mathrm{dist}(f(d, \theta), f(d', \theta')) &= \mathrm{dist}((\Tr(d, \theta), G(d, \theta)), (\Tr(d', \theta'), G(d', \theta'))) \\
	        &= \mc{W}_1(\Tr(d, \theta), \Tr(d', \theta')) + \lVert G(d, \theta) - G(d', \theta')\rVert_2.
	    \end{align*}
	       The $\eps$-joint sensitivity of the transition map yields
    \begin{equation}
        \label{sensitivity}
         \mc{W}_1(\Tr(d, \theta), \Tr(d', \theta')) \leq \eps \mc{W}_1(d, d') + \eps \lVert \theta - \theta' \rVert_2.
    \end{equation}
	 We will show that
	    \begin{equation}
	    \label{g_eps}
	        \lVert G(d, \theta) - G(d', \theta')\rVert_2 \leq \eps \frac{\beta}{\gamma}\mc{W}_1(d, d') +  \eps \frac{\beta}{\gamma} \lVert \theta - \theta' \rVert_2.
	    \end{equation}
	    Combining equations \eqref{sensitivity} and \eqref{g_eps} will conclude the proof.  
	    We obtain \eqref{g_eps} by using \Lem{perdomo} with distributions $\Tr(d, \theta)$ and $\Tr(d', \theta')$ together with the sensitivity of the transition map. This gives
	    \begin{align*}
	          \lVert G(d, \theta) - G(d', \theta')\rVert_2 \leq \frac{\beta}{\gamma}(\mc{W}_1(\Tr(d, \theta), \Tr(d', \theta')))
	          \leq \frac{\beta}{\gamma}(\eps \mc{W}_1(d, d') + \eps \norm{\theta-\theta'}).
	    \end{align*}
\end{proof}

\subsection{Proof of \Thm{finite-sample}}

In this section, we show \Thm{finite-sample} on the performance of RERM. In particular, we show that for iterates $(d_{t-1}, \theta_t)$ of RERM it holds that 
\begin{align*}
    \mathrm{dist}((d_{t-1}, \theta_t), ( d_\ST, \theta_\ST)) \leq \delta \text{ for all } t \geq  \Bigl(1 - 2\eps\Bigl(1+\frac{\beta}{\gamma}\Bigr) \Bigr) \log \Bigl( \frac{\mathrm{dist}((d_0, \theta_1), ( d_\ST, \theta_\ST))}{\delta}  \Bigr).
\end{align*}

\begin{proof}[Proof of \Thm{finite-sample}]
Given a distribution $d$, let $d^{(n)}$ denote the empirical distribution over the $n$ samples drawn from $d$. Let $\widetilde{G}(d) := G(d^{(n)})$. Define the RERM Map $\widetilde{f}(d, \theta) = (\Tr(d, \theta), \widetilde{G}(\Tr(d, \theta))$. Our analysis relies on the following assumption. 
\begin{assumption}\label{assumption:big_sample}
For each timestep $t \geq 1$, it holds that $\mc{W}_1(d_t^{(n_t)}, d_t) \leq  \eps\Bigl(1 + \frac{\gamma}{\beta}\Bigr) \delta$. 
\end{assumption}

Given \Assumption{big_sample}, we show that one of the following holds for each $t \geq 1$:

\begin{enumerate}
    \item If the iterate $(d_{t-1}, \theta_{t})$ is at distance at least $\delta$ from $(d_\ST, \theta_\ST)$, then the distance of the next iterate $(d_{t}, \theta_{t+1})$ to $(d_\ST, \theta_\ST)$ will contract by a factor of at least $2\eps\Bigl( 1 + \frac{\beta}{\gamma}\Bigr)$. 
    \item If the iterate $(d_{t-1}, \theta_{t})$ is within a ball of radius $\delta$ of $(d_\ST, \theta_\ST)$, then the next iterate will also be within this ball. 
\end{enumerate}

By Theorem 2 of \cite{FournierG15}, if $n_t = O\Bigl( \frac{1}{(\eps(1+\frac{\gamma}{\beta})\delta)^m} \log(t/p) \Bigr)$, then $\mc{W}_1(d_t^{(n_t)}, d_t) \geq  \eps\Bigl(1 + \frac{\gamma}{\beta}\Bigr) \delta$ with probability at most $\frac{6p}{\pi^2 t^2}$. By a Union Bound over all $t$, we obtain that \Assumption{big_sample} holds with probability at least $1 - \sum_{t=1}^{\infty} \frac{6p}{\pi^2 t^2} = 1 - p $.

We start by showing Item 1 is true under \Assumption{big_sample}. Suppose $\mathrm{dist}((d, \theta), ( d_\ST, \theta_\ST)) \geq \delta$. Then
\begin{align}
   &\mathrm{dist} (\widetilde{f}(d, \theta) , ( d_\ST, \theta_\ST)) \nonumber \\
   &= \mc{W}_1(\Tr(d, \theta), d_\ST) + \lVert \widetilde{G}(\Tr(d, \theta)) - \theta_\ST \rVert_2 \nonumber \\
   &\leq \mc{W}_1(\Tr(d, \theta), \Tr(d_\ST, \theta_\ST)) + \lVert \widetilde{G}(\Tr(d, \theta)) - G(\Tr(d, \theta)) \rVert_2 \label{eq:triangle}
   +  \norm{ G(\Tr(d, \theta)) - G(\Tr(d_\ST, \theta_\ST)) }\\
   &\leq \eps \mc{W}_1(d, d_\ST) + \eps \norm{\theta - \theta_\ST} + \frac{\beta}{\gamma}\mc{W}_1(\Tr(d, \theta)^{(n)}, \Tr(d, \theta)) + \frac{\beta}{\gamma}\mc{W}_1(\Tr(d, \theta), \Tr(d_\ST, \theta_\ST)) \label{eq:many_bounds} \\
   &\leq \eps \mc{W}_1(d, d_\ST) + \eps \norm{\theta - \theta_\ST} + \frac{\beta}{\gamma} \eps\Bigl(1 + \frac{\gamma}{\beta}\Bigr) \delta + \frac{\beta}{\gamma}( \eps \mc{W}_1(d, d_\ST) + \eps \norm{\theta - \theta_\ST}) \label{eq:assumption}\\
   &= \eps\Bigl(1+\frac{\beta}{\gamma}\Bigr)\mathrm{dist}((d, \theta), (d_\ST, \theta_\ST)) + \eps\Bigl(1+\frac{\beta}{\gamma}\Bigr)\delta \nonumber \\
   &\leq 2\eps\Bigl(1+\frac{\beta}{\gamma}\Bigr)\mathrm{dist}((d, \theta), (d_\ST, \theta_\ST)) \label{eq:delta}.
\end{align}
In \Eqn{triangle} we use the triangle inequality. \Eqn{many_bounds} follows by applying the $\eps$-Lipschitzness of the transition map to bound the first term and \Lem{perdomo} to bound the second and third term. In \Eqn{assumption}, we use \Assumption{big_sample} to bound the second term and $\eps$-Lipschitzness to bound the third term. Finally, in \Eqn{delta}, we use the assumption that $\mathrm{dist}((d, \theta), (d_\ST, \theta_\ST)) \geq \delta$. Therefore, Item 1 holds.

To show Item 2, suppose that $\mathrm{dist}((d, \theta), ( d_\ST, \theta_\ST)) < \delta$. We show that $\mathrm{dist} (\widetilde{f}(d, \theta) , ( d_\ST, \theta_\ST)) < \delta$. From the previous argument, under \Assumption{big_sample}, it holds
\begin{align*}
    \mathrm{dist} (\widetilde{f}(d, \theta) , ( d_\ST, \theta_\ST)) \leq \eps\Bigl(1+\frac{\beta}{\gamma}\Bigr)\mathrm{dist}((d, \theta), (d_\ST, \theta_\ST)) + \eps\Bigl(1+\frac{\beta}{\gamma}\Bigr)\delta \leq  2\eps\Bigl(1+\frac{\beta}{\gamma}\Bigr)\delta.
\end{align*}
From the assumption that $2\eps\Bigl(1+\frac{\beta}{\gamma}\Bigr) < 1$ we obtain Item 2.

It remains to show that under \Assumption{big_sample} the iterates $(d_{t-1}, \theta_t)$ will reach a ball of radius $\delta$ around $(d_{\ST}, \theta_{\ST})$ for $t \geq \Bigl(1 - 2\eps\Bigl(1+\frac{\beta}{\gamma}\Bigr) \Bigr) \log \Bigl( \frac{\mathrm{dist}((d_0, \theta_1), ( d_\ST, \theta_\ST))}{\delta}  \Bigr) $. Suppose the assumption on $t$ holds. Furthermore, suppose that none of the iterates $(d_0, \theta_1), \dots, (d_{t-2}, \theta_{t-1})$ are within a radius $\delta$ of $(d_\ST, \theta_\ST)$. By Assumption 1, Item 1, it holds
\begin{align*}
    \mathrm{dist}((d_{t-1}, \theta_t), ( d_\ST, \theta_\ST))
    &\leq \Bigl(2\eps\Bigl(1+\frac{\beta}{\gamma}\Bigr)\Bigr)^t \mathrm{dist}((d_0, \theta_1), ( d_\ST, \theta_\ST))  \\
    &\leq \exp\Bigl( -t\Bigl( 1 -  2\eps\Bigl(1+\frac{\beta}{\gamma}\Bigr)\Bigr)\Bigr)\mathrm{dist}((d_0, \theta_1), ( d_\ST, \theta_\ST)) \\
    &\leq \delta.
\end{align*}
If one of the iterates $(d_0, \theta_1), \dots, (d_{t-2}, \theta_{t-1})$ is within a radius $\delta$ of $(d_\ST, \theta_\ST)$, then by Assumption 1, Item 2, all consecutive iterates will also be within a radius $\delta$ of a stable pair. Since Assumption 1 holds with probability $1-p$, this concludes the proof. 
\end{proof}

\subsection{Proof of \Thm{rrm_optimality}}

\begin{proof}[Proof of \Thm{rrm_optimality}] Our proof is similar to an argument of \cite{perdomo2020}.
Let $\theta_\OPT$ be an optimal classifier and let $d_\OPT$ be its corresponding fixed point distribution. Let $\theta_\ST$ be a stable classifier, with corresponding fixed point distribution $d_\ST$. By the definitions of optimality and stability,
\begin{equation}\label{eq:inequality}
   \underset{z \sim d_\OPT}{\E} \ell(z; \theta_\OPT) \leq 
   \underset{z \sim d_\ST}{\E} \ell(z; \theta_\ST) \leq 
   \underset{z \sim d_\ST}{\E} \ell(z; \theta_\OPT) \:.
\end{equation}
We first show that
\begin{equation}
\label{eq:po_ps}
    \underset{z \sim d_\ST}{\E} \ell(z; \theta_\OPT) -  \underset{z \sim d_\ST}{\E} \ell(z; \theta_\ST) \geq \frac{\gamma}{2} \lVert \theta_\OPT - \theta_\ST \rVert_2^2\:. 
\end{equation}
By the strong convexity of the loss function, for all $z \in \mc{Z}$ it holds
\[
\ell(z;\theta_\OPT) \geq \ell(z;\theta_\ST) + \nabla_\theta \ell(z;\theta_\ST)^\top (\theta_\OPT - \theta_\ST) + \frac{\gamma}{2}\lVert \theta_\OPT - \theta_\ST \rVert_2^2\:.
\]
As a result,
\[
\underset{z \sim d_\ST}{\E} \bigl[ \ell(z; \theta_\OPT) - \ell(z; \theta_\ST) \bigr] \geq \underset{z \sim d_\ST}{\E} \bigl[  \nabla_\theta \ell(z;\theta_\ST)^\top (\theta_\OPT - \theta_\ST) \bigr] + \frac{\gamma}{2}\lVert \theta_\OPT - \theta_\ST \rVert_2^2\:.
\]
Since $\theta_\ST$ minimizes the value of $\ell$ over the distribution $d_\ST$, \Lem{bubeck} implies
\[
\underset{z \sim d_\ST}{\E} \bigl[  \nabla_\theta \ell(z;\theta_\ST)^\top (\theta_\OPT - \theta_\ST) \bigr]  \geq 0.
\]
Therefore, \Eqn{po_ps} holds. On the other hand, since the loss is $L_z$-Lipschitz in $z$, by \Lem{sensitivity},
\[
\underset{z \sim d_\ST}{\E} \ell(z; \theta_\OPT) -  \underset{z \sim d_\OPT}{\E} \ell(z; \theta_\OPT) \leq L_z \mc{W}_1(d_\ST, d_\OPT) \leq \frac{L_z\eps}{1-\eps}\lVert \theta_\OPT - \theta_\ST \rVert_2\:.
\]
If $\frac{\eps}{1-\eps} < \frac{\gamma \lVert \theta_\OPT - \theta_\ST \rVert_2}{2L_z}$ then $\frac{L_z\eps}{1-\eps} \lVert \theta_\OPT - \theta_\ST \rVert_2 \leq \frac{\gamma}{2}\lVert \theta_\OPT - \theta_\ST \rVert_2^2$. This would imply that 
\[
\underset{z \sim d_\ST}{\E} \ell(z; \theta_\OPT) -  \underset{z \sim d_\OPT}{\E} \ell(z; \theta_\OPT) \leq   \underset{z \sim d_\ST}{\E} \ell(z; \theta_\OPT) -  \underset{z \sim d_\ST}{\E} \ell(z; \theta_\ST), 
\]
which contradicts \Eqn{inequality}. Therefore, $\frac{\eps}{1-\eps} \geq \frac{\gamma \lVert \theta_\OPT - \theta_\ST \rVert_2}{2L_z}$, as desired.
\end{proof}

\subsection{Proof of \Thm{kGRS}}

In this section, we show \Thm{kGRS} on the performance of RRM and $k$-Delayed RRM for the setting of \Ex{kGRS}.

\begin{proof}[Proof of \Thm{kGRS}] 

 \textbf{$k$-Delayed RRM.} Let $(d_{kt-1}, \theta_{kt})$ be the iterates of \Alg{kDelayed} at timestep $kt$ for $t \geq 1$. Since $kt$ is a multiple of $k$, then $\theta_{kt} = G(d_{kt-1})$ and all components $d_{kt}^{|j}$, $j \in [k]$ of the mixture distribution $d_{kt-1}$ are identical. Therefore $\theta_{kt} = G(d_{kt-1}^{|k})$. 
Additionally, $d_{k(t+1)-1}^{|k} = \D(\theta_{kt})$. Therefore, the iterates $(d_{kt-1}^{|k}, \theta_{kt})$, where $t \geq 1$, correspond to the iterates of RRM in the stateless setting of \cite{perdomo2020}. By Theorem 3.5 of \cite{perdomo2020}, the iterates $\theta_{kt}$ converge to $\theta_\ST$ at rate $\Bigl(1-\eps\frac{\beta}{\gamma}\Bigr)^t$. In turn, it follows that the iterates of \Alg{kDelayed} converge to $\theta_\ST$ at rate $k\Bigl(1-\eps\frac{\beta}{\gamma}\Bigr)^t$. 

 \textbf{RRM.} Let $\theta_\ST$ denote a stable classifier for $\D(\theta)$, and let $d_\ST = \D(\theta_\ST)$. Let $\mathbf{d}_{\ST} = (d_\ST, \dots, d_\ST)$ be the uniform mixture on $k$ identical components $d_\ST$. Then, $(\mathbf{d}_{\ST}, \theta_\ST)$ is a stable point for $\Tr$ because $\Tr(\mathbf{d}_{\ST}, \theta_\ST) = (\D(\theta_\ST), d_\ST, \dots, d_\ST) = \mathbf{d}_{\ST}$, and $\theta_\ST$ is the classifier that minimizes loss on $\mathbf{d}_{\ST}$. Let $(d_{t-1}, \theta_t)$ denote the iterates of RRM in the setting of \Ex{kGRS} and suppose $t \geq k$.  Then,
\begin{align*}
    \norm{\theta_{t} - \theta_\ST} = 
    \norm{G(d_{t-1}) - G(\theta_\ST)} &\leq \frac{\beta}{\gamma} \mc{W}_1(d_{t-1}, \mathbf{d}_{\ST}) \\
    &\leq \frac{\beta}{\gamma} \frac{1}{k}\sum_{i=1}^{k} \mc{W}_1 (d_{t-1}^{|i}, d_\ST) \\
    &= \frac{\beta}{\gamma} \frac{1}{k}\sum_{i=1}^{k} \mc{W}_1 (\D(\theta_{t-i}), \D(\theta_\ST)) \\
    &\leq \frac{\beta}{\gamma} \frac{1}{k}\sum_{i=1}^{k} \norm{\theta_{t-i} - \theta_S},
\end{align*}
where the first inequality follows by \Lem{perdomo}, the second inequality follows by the definition of $d_{t-1}$ as a uniform mixture on $k$ components, and the third inequality follows by the $\eps$-sensitivity of the map $\D(\cdot)$. 

Let $\delta_t =  \norm{ \theta_{t} - \theta_\ST}$. To find the rate of convergence of the sequence $\delta_t, t\geq 1$, it suffices to find the rate of convergence for the sequence that satisfies the following linear recurrence
\begin{align*}
    \delta_{t} = \frac{\eps \beta}{\gamma} \cdot \frac{1}{k} (\delta_{t-1} + \dots + \delta_{t-k}).
\end{align*}

The sequence decays exponentially with a rate the depends on $\eps, \beta, \gamma, k$. We provide a simpler analysis, using the inequality
\begin{align*}
    \delta_{t} \leq \frac{\eps \beta }{\gamma} \max_{i \in [k]} \delta_{t-i}.
\end{align*}

 For $j \geq 0$, let $i_j^* = \arg\max_{i \in [k]} \delta_{jk+i}$. Define $\tilde{\theta}_j = \theta_{i_j^*}$. By the above argument, $\delta_{jk+1} \leq  \frac{\eps \beta }{\gamma}  \max_{i \in [k]} \delta_{jk+1-i} = \max_{\ell \in [k]} \delta_{(j-1)k + \ell} =   \frac{\eps \beta }{\gamma} \norm{\tilde{\theta}_{j-1} - \theta_\ST}$. 

 Now, $\delta_{jk+2} \leq  \frac{\eps \beta }{\gamma} \max_{i \in [k]} \delta_{jk+2-i} \leq  \frac{\eps \beta }{\gamma} \max_{i \in [k+1]} \delta_{jk+2-i} \leq  \frac{\eps \beta }{\gamma} \max \{\delta_{jk+1}, \max_{\ell \in [k]} \delta_{(j-1)k + \ell}\} \leq  \frac{\eps \beta }{\gamma} \norm{\tilde{\theta}_{j-1} - \theta_\ST}$. 

 By an inductive argument, we can show that for all $i \in [k]$, 
 $\delta_{jk+i} \leq \frac{\eps \beta}{\gamma }\norm{\tilde{\theta}_{j-1} - \theta_\ST}$. It follows that $\norm{\tilde{\theta}_{j} - \theta_\ST} \leq \frac{\eps\beta}{\gamma} \norm{\tilde{\theta}_{j-1} - \theta_\ST}$. Therefore, the sequence of classifiers $\tilde{\theta}_1, \tilde{\theta}_2, \dots $ converges to $\theta_\ST$ at a rate of $\Bigl(1-\eps\frac{\beta}{\gamma}\Bigr)^t$ for $\eps < \frac{\gamma}{\beta}$. It follows that the sequence $\theta_1, \theta_2, \dots$ converges to $\theta_\ST$ at a rate $k\Bigl(1-\eps\frac{\beta}{\gamma}\Bigr)^t$.
 \end{proof}

\section{SIMULATION}

\subsection{Additional Details}\label{app:simulation}

We simulate a credit score system using the dataset \emph{GiveMeSomeCredit} \cite{givemesomecredit} from Kaggle. Before giving a loan to an applicant, a bank tries to predict whether the individual will experience financial distress in the next two years. Hence, individuals prefer a \textit{negative} classification. The prediction is based on 11 biographic and financial history features included in the dataset. 

In the rest of this section we describe the deterministic ``strategic response function'' $\D(\theta)$ of the $k$ Groups Respond Slowly model (\Ex{GDR}) used in the simulation. Recall that the distribution of the group $j$ at round $t \geq j$ is $\D(\theta_{t-j+1})$.

In the simulation, we assume that the world population is finite, consisting of exactly the individuals in the dataset.
There are 18,357 individuals (data points).
The distribution is uniform over these individuals, who may change their features strategically, resulting in a new distribution.
The original dataset serves as both the initial distribution of each group ($d_0^{|j}$) and the ``baseline'' distribution $d_{\mathrm{BL}}$, from which modifications are made.

The best response of an
individual $(x, y) \in d_{\mathrm{BL}}$ is
\[
x_{\mathrm{BR}}(\theta) \xleftarrow \arg\max_{x'} u(x', \theta) - c(x', x),
\]
where $u$ is the utility function and $c$ is the cost function. The family of classifiers $\Theta$ is logistic regression. We use
\[
u(x) = - \langle \theta, x \rangle,
\]
because a negative value for the utility translates into the more favorable negative prediction. We consider a quadratic cost for feature updates:
\[
c(x', x) = \frac{1}{2\varepsilon} \lVert x' - x \rVert^2_2.
\]
In our experiments, the main parameter we adjust is the sensitivity $\varepsilon$, which controls the strength of strategic response. Additionally, we assume that the individual can change only a subset $S$ of her features, which we call the \textit{strategic features}.
Let $x^S$ be the restriction of $x$ to $S$. Solving the maximization problem of the individual leads to the response
\[
x^S_{\mathrm{BR}}(\theta) = x^S - \eps \theta^S.
\]  
The rest of the features remain unchanged. With that, we can define the strategic response function $\D(\theta)$ as
\[
\D(\theta) = \mathrm{Uniform}\left(\{(x_{\mathrm{BR}}(\theta), y) | (x, y) \in d_{\mathrm{BL}}\}\right). 
\]

\subsection{Different Initial  Distributions}\label{app:sim-init}
\begin{figure}
     \centering
     \begin{subfigure}[b]{\textwidth}
         \centering
         \includegraphics[width=\textwidth]{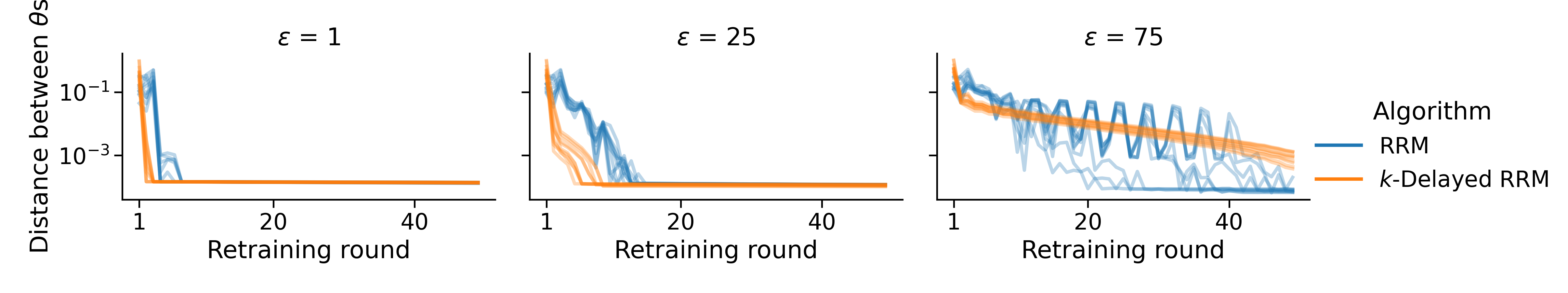}
         \caption{ Convergence of RRM and $k$-Delayed RRM for varying values of $\eps$. The horizontal axis shows the number of \emph{retraining rounds} and the vertical axis shows the distance between successively trained $\theta$s.}
     \end{subfigure}
     \vfill
     \begin{subfigure}[b]{\textwidth}
         \centering
         \includegraphics[width=\textwidth]{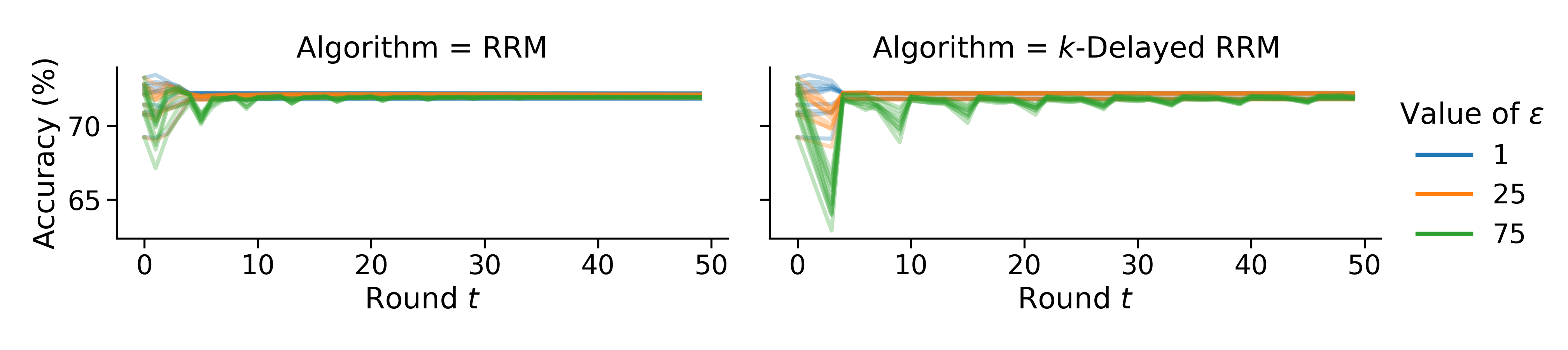}
         \caption{ Accuracy of RRM and k-Delayed RRM for different $\eps$. The horizontal axis shows the number of \emph{rounds} and the vertical axis shows the accuracy of the published model after the strategic response.}
     \end{subfigure}
     \vfill
     \begin{subfigure}[b]{\textwidth}
         \centering
     \includegraphics[width=\textwidth]{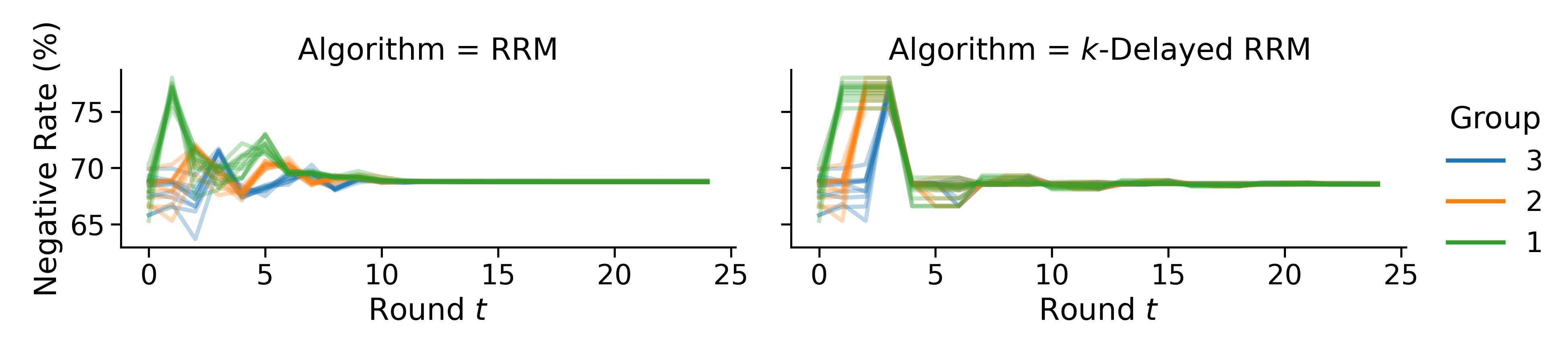}
         \caption{Negative rate of RRM and k-Delayed RRM for $\eps = 25$. The horizontal axis shows the number of \emph{rounds} and the vertical axis shows the negative rate of the published model after the strategic response.}
         \label{fig:five over x}
     \end{subfigure}
        \caption{Comparison of RRM and $k$-Delayed RRM for for $k = 3$ over three different metrics. We run the experiments from the main text ten times, but with different initial distribution for each group. Each line represents the results of an experiment with a random different group-initialization.}
        \label{fig:sim-int}
\end{figure}

We repeat the experiments from Section~\ref{sec:simulation} ten times but with different initial distribution for each group. The initial distribution of group $j \in [k]$, namely $d_0^{|j}$, is generated by sampling \emph{with replacement} $N=18,357$ individuals from the \emph{GiveMeSomeCredit} dataset. Figure~\ref{fig:sim-int} shows that we our findings from the setting where all groups have identical initial distributions also hold for this setup, with the exception of the slower convergence of RRM for some of the initializations when $\eps=75$

\end{document}